\documentclass{article}

\usepackage{arxiv}
\usepackage[numbers,sort&compress]{natbib}

\usepackage[utf8]{inputenc} 
\usepackage[T1]{fontenc}    
\usepackage{hyperref}       
\usepackage{url}   
\usepackage{algorithm}
\usepackage{booktabs}       
\usepackage{amsfonts}       
\usepackage{nicefrac}       
\usepackage{microtype}      
\usepackage{lipsum}
\usepackage{graphicx}
\usepackage{amsmath} 
\usepackage{amssymb} 
\usepackage{caption}
\usepackage{pgfplots}
\usepackage{tikz}
\usepackage{enumitem}
\usepackage{subcaption}
\usepackage{makecell} 
\usepackage{graphicx}

\usetikzlibrary{arrows.meta, positioning}
\usetikzlibrary{shapes.geometric, arrows.meta, positioning}

\usetikzlibrary{positioning, calc,decorations.pathreplacing}

\usepackage{pgfplotstable}
\pgfplotsset{compat=1.18}

\usepackage[noend]{algpseudocode} 

\usepackage[utf8]{inputenc}

\usepackage{algpseudocode} 

\graphicspath{ {./images/} }

\title{High Cursive Complex Character Recognition using GAN External Classifier \\[0.5em]
\large Proceedings of the 2nd International Conference on Computing Advancements 2022}

\author{
\textbf{S M Rafiuddin} \\
Department of Computer Science and Engineering \\
University of Asia Pacific \\
\texttt{rifat.cse@uap-bd.edu}
}

\begin{document}
\maketitle

\begin{abstract}
Handwritten characters can be trickier to classify with the increase of their complex and cursive nature in comparison with the characters that are simple and non-cursive. We present an external classifier along with a Generative Adversarial Network that has the capability of classifying highly cursive and complex characters. The generator network will generate fake handwritten character images, and these generated images are used to augment training data after adding some adversarially perturbed noise and getting a confidence metric above the threshold value with the discriminator network. This result shows that the accuracy of the convolutional neural networks degrades with the increase of complexity of the character images, but our proposed model \textit{i.e.} ADA-GAN is more robust and sophisticated with high cursive as well as complex characters. 
\end{abstract}


\maketitle

\section{Introduction}
Deep learning techniques are gaining interest across all kinds of image classification procedures. As deep learning models have a reputation for their data-hungry nature, it is not always possible to achieve the level of accuracy demanded. Creating a large dataset is very expensive and often it is not possible to gather such a huge amount of data for a specific task. 

In this paper, we focused on the classification tasks on handwritten characters which are more complex than the typical English handwritten characters \textit{e.g.}, Chinese, Japanese, Korean, Bengali, etc. \cite{li2020deep} \cite{jalali2020high} \cite{hasan2020bangla} \cite{rabby2020borno} \cite{azad2020bangla} \cite{gan2020compressing} \cite{oktaviani2020optical}. We choose Bengali handwritten characters as they can be both simple in structure and as well as cursive and complex. So, as Bengali handwritten characters can be both simple and complex in structure \cite{rabby2018ekush}, applying the classification models to the Bengali handwritten character dataset would give us a clear indication of how the model would perform on both simple handwritten characters like English, and complex handwritten characters like Chinese and Korean.

\begin{figure}[!htbp]
\centerline{\includegraphics[width=5cm]{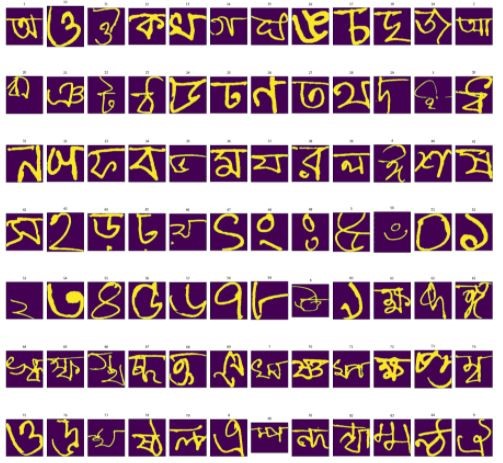}}
\vspace*{8pt}
\caption{Eighty Four (84) distinct images of BanglaLekha Isolated Handwritten Character Dataset. This dataset has simple characters as well as highly cursive complex characters.}
\end{figure}

In our paper, several deep learning models \textit{e.g.}, 3D CNN \cite{ji20123d}, ResNet50 \cite{he2016deep}, MobileNet  \cite{howard2017mobilenets}, and InceptionV3 \cite{chen2019modified} have been applied to the BanglaLekha Isolated Handwritten Character dataset \cite{biswas2017banglalekha} as well as the MNIST dataset \cite{lecun1998mnist}. The result shows that all models got almost 99\% accuracy for the MNIST dataset, but, the accuracy varies from model to model for the BanglaLekha Isolated Handwritten Character dataset. So, it is very clear that, as the complexity and cursiveness grow, it is more difficult to classify handwritten characters for the models. The result also suggests that the InceptionV3 model is more robust and sophisticated to classify more complex handwritten characters among the other models mentioned.

In our paper, we tried a different approach to enhance the accuracy even further to classify the complex and high-cursive characters. We used Generative Adversarial Network (GAN) \cite{goodfellow2014generative} along with an external classifier to generate more data, and added some adversarial noise to these generated data. Then, these generated data with adversarial noise are fed into the external classifier with a confidence metric. If the classifier confidence metric is above the threshold value, then the data, \textit{i.e.}, character images are augmented to the training dataset. As we know, a GAN has two special Artificial Neural Networks \textit{i.e.} Generator and Discriminator, and these two networks play a combative role between themselves and make the fake image generation more trustworthy to the discriminator. We added an external classifier to enhance the accuracy with these augmented images.

This proposed model has better accuracy in high cursive as well as simple characters. This model will also be beneficial for any low-resource training data condition.

This paper begins with a discussion on the topics related to the prospects and challenges of high-cursive complex character recognition, data augmentation using Generative Adversarial Networks, and how it is beneficial to augment noise to create adversarially perturbed noise to the images regarding image classification. Then, in the subsequent sections, we discussed our proposed method, dataset, and experimental results.

\section{Related Work}

\subsection{High Cursive Complex Handwritten Character Recognition}

There are many languages where the characters are very complex compared to the typical English handwritten characters. Many ‘state-of-the-art’ deep learning model has been applied to the classical MNIST dataset \cite{lecun1998mnist} to determine the classification accuracy. But some issues arise when the characters are more complex \textit{e.g.}, Chinese handwritten characters \cite{li2020deep} \cite{gan2020compressing}, Korean traditional handwritten characters \cite{jalali2020high} \cite{oktaviani2020optical}, Bengali Handwritten Characters \cite{hasan2020bangla} \cite{rabby2020borno} \cite{azad2020bangla}, and so on. Several papers show that as the complexity of the structures of the handwritten characters grows, the performance of the deep learning model degrades. This degradation is due to the convoluted edge, characters that look different but belong to the same class, and characters that look similar but belong to different classes \cite{hasan2020bangla}.

\begin{figure}[!htbp]
\centerline{\includegraphics[width=5cm]{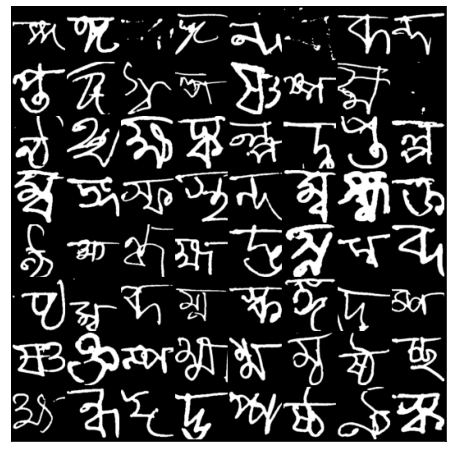}}
\vspace*{8pt}
\caption{A sample of High Cursive Complex Character from BanglaLekha Isolated Character Dataset}
\end{figure}

\subsection{Artificial Neural Network Models for Character Recognition}

Many different deep learning models have been tried to classify the handwritten characters, such as DenseNet \cite{alom2018handwritten}, VGGNet \cite{mudhsh2017arabic}, ResNet50 \cite{he2016deep}, and Inception model \cite{chen2019modified} to classify high cursive complex characters \cite{hasan2020bangla}. But it has been observed that as the complexity of high cursive complex handwritten character grows, the Inception model \cite{szegedy2015going} \cite{szegedy2016rethinking} \cite{szegedy2017inception} gives robust accuracy compared to other models \cite{hasan2020bangla}. In this paper, we also observed that, as the handwritten character grows more complex compared to single numeric characters, the Inception model is more robust and gives higher accuracy to classification as it has more depth and a wider network to classify the convoluted edge of complex handwritten characters.

\subsection{Classification using GAN}

There has been an ample amount of classification work done in the recent past using generative adversarial networks. The generator model and the discriminator model try to combat each other to get more accurate, as the generator model tries to fool the discriminator model by creating more convincing data, and at the same time, the discriminator model tries to determine whether the image of the generator seems real or fake. So, this technique is useful for multiclass classification where the discriminator model can be used to classify images into various classes using the Softmax function, where the training sample is low \cite{ali2019mfc} \cite{haque2020ec}.   

\subsection{Data Augmentation Approach using GAN}

Since the Generative Adversarial Networks can generate meaningful artificial data and hence this technique has been used to augment the dataset \cite{haque2020ec} \cite{golany2020improving} \cite{frid2018synthetic} \cite{chaudhari2019data}. As it creates from the training data distributions, this technique is used as an efficient method. It is better than the traditional data augmentation like rotation and flipping methods.

\subsection{Neural Structured Learning}

Goodfellow \textit{et. al.} \cite{goodfellow2014explaining} showed that the classification accuracy of a base model degrades abruptly if the input images are perturbed adversarially, by augmenting some random noise into it. So, if we incorporate an image with adversarially perturbed data or adversarial image examples to train our base model, then the classification model will not degrade its performance by that much amount \textit{i.e.} it will degrade a very less amount compared to the previous. So, in our proposed model, our approach is to perturb images by some random noise generated by GAN and use a confidence metric to decide which data has the capability and quality to augment into the training image dataset after adversarial random noise perturbation.

\section{Method}

\subsection{Adversarial Data Generation}

A Generative Adversarial Network mainly has two distinct Artificial Neural Networks \cite{goodfellow2014generative} \textit{i.e.} the Generator and the Discriminator. The Generator Neural Network generates the images based on the approximation of the training image dataset distribution. At the same time, the Discriminator Neural Network tries to determine whether the generated images are real images or fake images. These coupled procedures go on as the Generator model tries to generate more qualitatively accurate images and the Discriminator model tries to improve the prediction of whether the images are fake or not, these procedures combined, eventually improving the data generation procedure. The objective of the GAN network is \cite{goodfellow2014generative}—

$${\mathbb{E}_x [log(D(x))]+\mathbb{E}_x[log(1-D(G(z)))]}$$

Here, ‘D’ is the Discriminator model and ‘G’ is the Generator model.\\
$\mathbb{E}_x$ = Expected value of all training data that fed into the model.\\
D(x) = Discriminator probability of confidence of the image ‘x’ is real.\\
$\mathbb{E}_z$ = Expected value of random training data that fed into the model.\\

\subsection{Augment Adversarial Noises to the generated images}
As the GAN architecture generates continuously improved fake images, there are some problems of the misclassification of the generated images. Goodfellow \textit{et. al.} \cite{goodfellow2014explaining} showed that, if images are adversarially perturbed with random noise, then the classifier often misclassifies the image with a high confidence rate. So, in this regard, the overall base model accuracy degrades significantly. In our proposed model, after generating the fake images from GAN networks, we added some adversarial random noise to the generated images before augmenting these qualitatively improved generated images to the training dataset. This procedure deals with the problem of degradation of athe ccuracy of the base classifier model more robustly.

\begin{figure}[!htbp]
\centerline{\includegraphics[width=5cm]{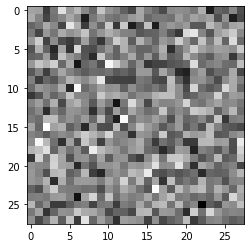}}
\vspace*{8pt}
\caption{Adversarial Noise added to the GAN-generated image.}
\end{figure}

\subsection{Augment Data using GAN}
After adding perturbed noise adversarially, images are augmented to the training dataset using a threshold value of 0.7. That is, after generating the images and adding adversarial noises to the images if the confidence value of $C(G(z))$ is greater than 0.8, then the images are augmented to the training dataset. 

\subsection{Classification with Adversarial Augmented Data}
Since the adversarial perturbed images are augmented to the training dataset, an external classifier is incorporated to classify the training data. As the augmented images are generated by the distribution of real training images with adversarial noise, it is more convenient for the external classifier to classify the testing dataset images with a wider variety. From the result, we can see that, for complex and high-cursive images of Bengali characters, our proposed model got better accuracy than the various state-of-the-art Convolutional Neural Network architectures.

As the ADA-GAN consists of three distinct Artificial Networks, namely the generator, the discriminator, and the classifier, their associated loss is defined as--

Generator loss, $\mathbb{L}_G$\cite{goodfellow2014generative}--
$$ log (1 - D(z)) $$

Discriminator loss, $\mathbb{L}_D$ \cite{goodfellow2014generative}--
$$ log (D((x)) + log (1 - D(G(z))) $$

And our modified external classifier loss, $\mathbb{L}_C$-
$$Cross Entropy (C (d), Y)  + Cross Entropy (C(G(z), Y))$$
Here, $d$ = Training dataset images. \\
$Y$ = Label of the training dataset images. \\
$C$ = External Classifier.

\subsection{Proposed Model}

The classifier is used which Artificial Neural Network architecture has the best accuracy with the prior training dataset. All images are rescaled to 28*28 pixels, and the batch size for all classifiers is 32. The softmax function has been used as the activation function of the external classifier. The algorithmic procedure for the proposed ADA-GAN model is as follows-- 

\begin{algorithm}
\caption{ADA-GAN}\label{alg:ada-gan}
\begin{algorithmic}[1]
\Require $d$ \Comment{Dataset images}
\Ensure the External classifier is trained on augmented data
\Statex

\State \textbf{3DCNN} $\gets \textsc{Train3DCNN}(d)$
\State \textbf{ResNet50} $\gets \textsc{TrainResNet50}(d)$
\State \textbf{MobileNet} $\gets \textsc{TrainMobileNet}(d)$
\State \textbf{InceptionV3} $\gets \textsc{TrainInceptionV3}(d)$

\State \textbf{VotedClassifier} $\gets \textsc{MaxAccuracy}(\textbf{3DCNN}, \textbf{ResNet50}, \textbf{MobileNet}, \textbf{InceptionV3})$

\State $d' \gets \textsc{GAN}(d)$
\If{$C(G(z)) \ge 0.8$}
    \State $d \gets d \cup d'$ \Comment{Augment with high-quality GAN samples}
\EndIf

\State \textbf{AugData} $\gets \textsc{AdversarialNoise}(d)$
\State \textbf{ExternalClassifier} $\gets \textsc{TrainExternal}(\textbf{VotedClassifier}, \textbf{AugData})$
\end{algorithmic}
\end{algorithm}

Here, three distinct Artificial Neural Networks are involved. Training images distribution is fed into the generator, which creates fake images. The fake images go into the discriminator, which predicts whether the images are real or fake. The predicted real images withthe  above confidence threshold are augmented with adversarial random noises to create adversarial perturbed images, which are again augmented into training image datasets to classify with an external classifier more efficient way.

Schematic diagram of ADA-GAN--

\begin{figure}[!htbp]
\centerline{\includegraphics[width=8.5cm]{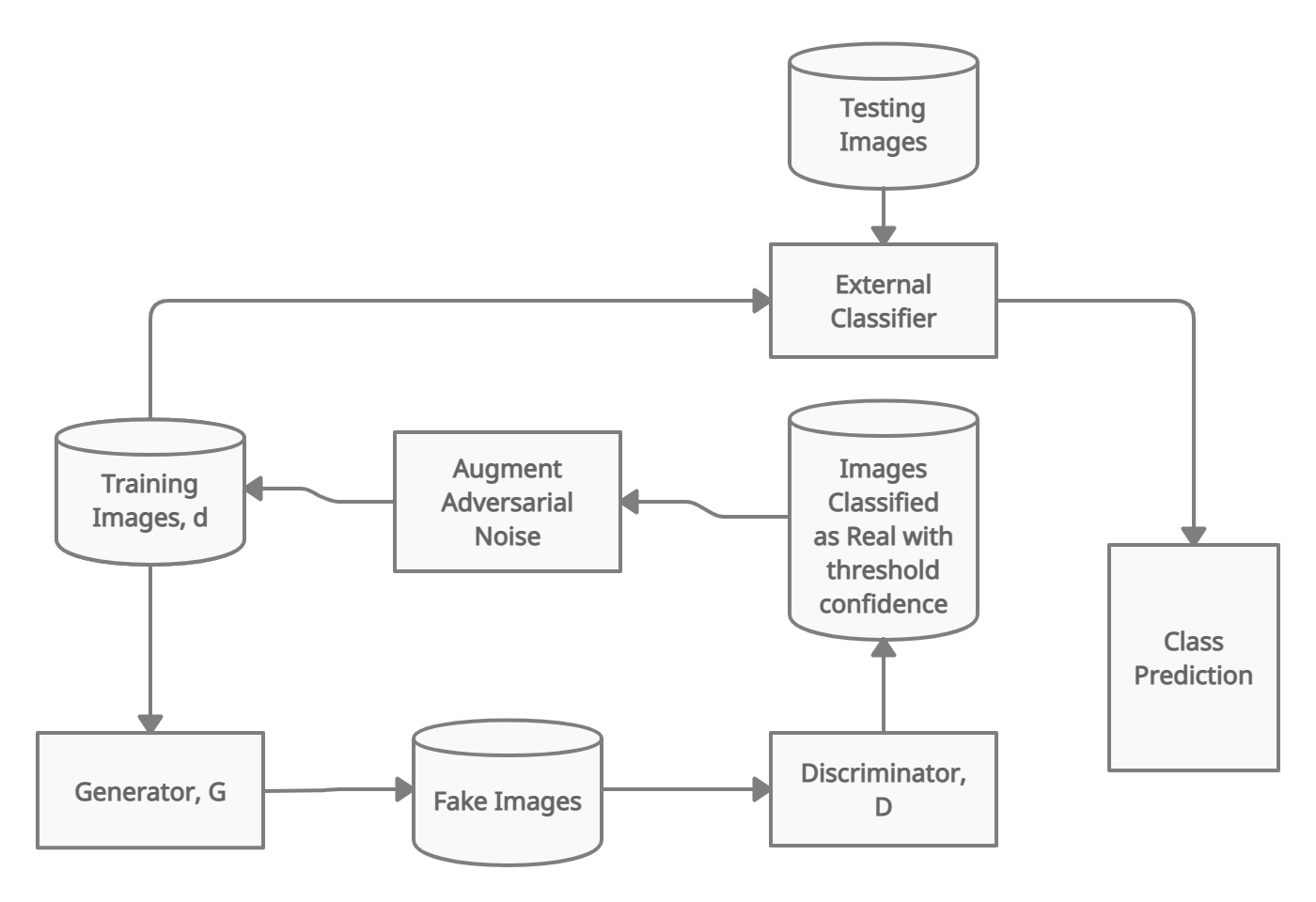}}
\vspace*{8pt}
\caption{The structure of ADA-GAN. }
\end{figure}

\section{Experiment and Result}

\subsection{Datasets}

\renewcommand{\theenumi}{\roman{enumi}}%
\begin{enumerate}
  \item \textbf{MNIST:} \\
MNIST is an English handwritten numeric character dataset widely used to measure the accuracy of various Machine Learning image classification models (Figure 5). It consists of 60,000 training image samples and 10,000 testing image samples. As the English numeric handwritten characters are simple in structure compared to the handwritten character in other languages like Chinese, Korean, and Bengali, it is easy for the classifier models to classify these images with a high accuracy rate of almost above 98\% \cite{lecun1998mnist}.
  \item \textbf{BanglaLekha Isolated Handwritten Character Dataset:}
 BanglaLekha Isolated Dataset is a collection of 1,66,105 Bengali Handwritten Character images within 84 distinct classes. Each class contains almost 2,000 images each. The dataset contains 11 vowel characters, 39 consonant characters, 10 numeric characters, and 24 complex and high-cursive characters \cite{biswas2017banglalekha}.  
\end{enumerate}

\subsection{Result}

From the result, we can see that, there is no such difference between the other classifier models and our external augmented classifier GAN model (ADA-GAN) when the characters are simple, \textit{i.e}, for numeric, as well as vowel and consonant characters. But when it comes to complex characters, our proposed model is more sophisticated and robust than the other classifier models (Table 2). Also, the overall performance for the entire dataset, ADA-GAN, has better accuracy than the others. 

From these results, we can assert that, when the complexity of the handwritten characters grows, augmenting images generated by GAN to the training dataset and perturbed by adding adversarial noise to the generated images, the classifier model can sustain the classification accuracy as complexity grows.

\begin{figure}[!htbp]
\centering
\begin{subfigure}{.4\textwidth}
  \centering
  \includegraphics[width=.9\linewidth]{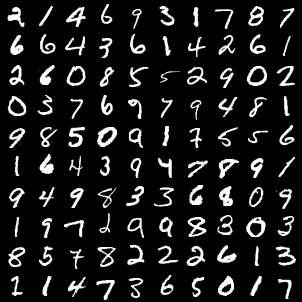}
  \caption{Sample of MNIST data.}
\end{subfigure}%
\hspace{1cm} 
\begin{subfigure}{.4\textwidth}
  \centering
  \includegraphics[width=.9\linewidth]{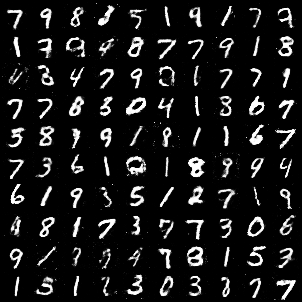}
  \caption{GAN-generated MNIST image after 300 epochs.}
\end{subfigure}
\caption{Real and fake sample images of the MNIST Dataset.}
\label{fig:MNIST}
\end{figure}

\begin{table}[ht]
\centering
\caption{Comparison of the accuracy of various models for 10 epochs on the MNIST dataset.}
\label{tab:commands}
\begin{tabular}{lcc}
\toprule
\textbf{Model} & \textbf{Training Accuracy} & \textbf{Testing Accuracy} \\
\midrule
3D CNN      & 99.13 & 98.18 \\
ResNet50    & 99.45 & 98.69 \\
MobileNet   & 99.78 & 99.13 \\
InceptionV3 & 99.65 & 99.51 \\
ADA-GAN     & 99.41 & 99.39 \\
\bottomrule
\end{tabular}
\end{table}

\begin{figure}[!htbp]
\centering
\includegraphics[width=5cm]{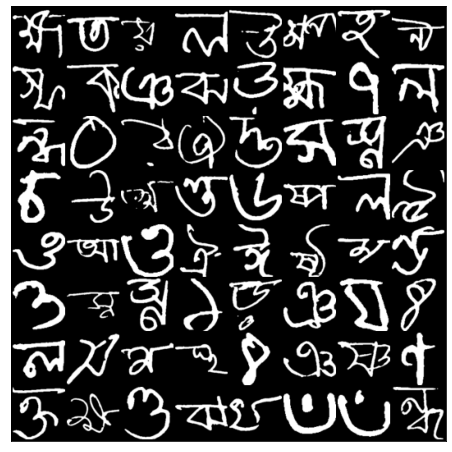}
\caption{Sample of BanglaLekha Isolated Data}
\end{figure}

\begin{figure}[!htbp]
\centering
\includegraphics[width=5cm]{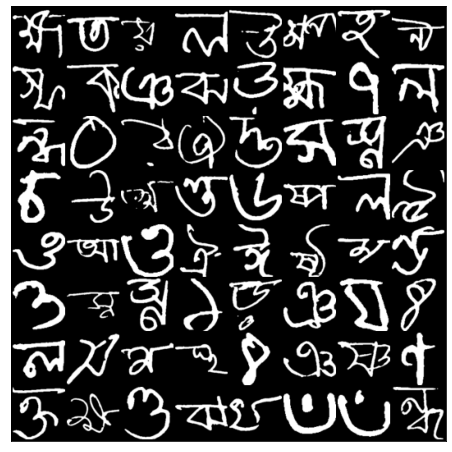}
\caption{Sample of BanglaLekha Isolated Data}
\end{figure}

\begin{figure*}[!htbp]
\centering
\begin{subfigure}{.3\textwidth}
  \centering
  \includegraphics[width=.8\linewidth]{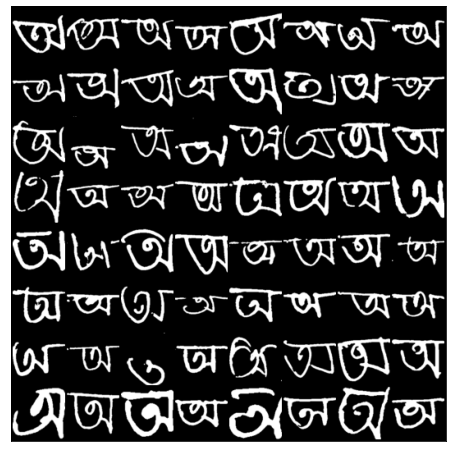}
  \caption{BanglaLekha Isolated Vowel Image of one class.}
\end{subfigure}%
\hfill
\begin{subfigure}{.3\textwidth}
  \centering
  \includegraphics[width=.8\linewidth]{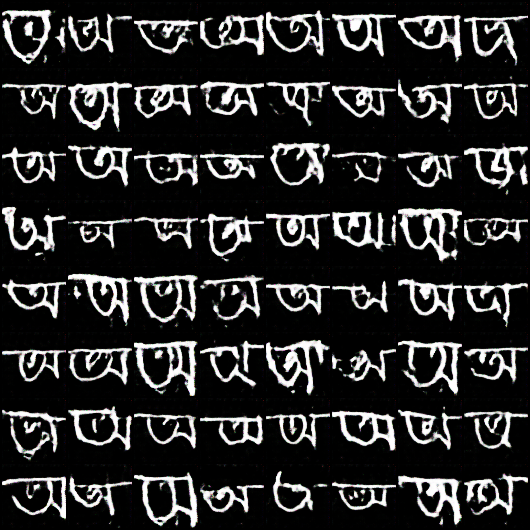}
  \caption{GAN-generated BanglaLekha Isolated Vowel Image of one class after 300 epochs.}
\end{subfigure}

\begin{subfigure}{.3\textwidth}
  \centering
  \includegraphics[width=.8\linewidth]{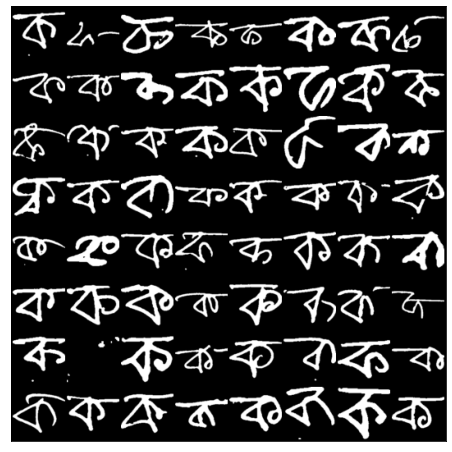}
  \caption{BanglaLekha Isolated Consonant Image of one class.}
\end{subfigure}%
\hfill
\begin{subfigure}{.3\textwidth}
  \centering
  \includegraphics[width=.8\linewidth]{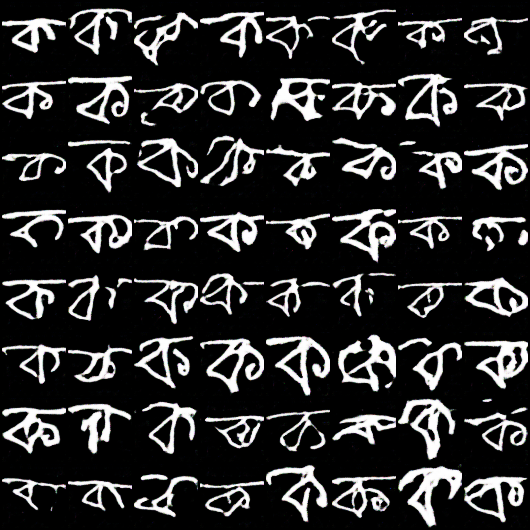}
  \caption{GAN-generated BanglaLekha Isolated Consonant Image of one class after 300 epochs.}
\end{subfigure}

\begin{subfigure}{.3\textwidth}
  \centering
  \includegraphics[width=.8\linewidth]{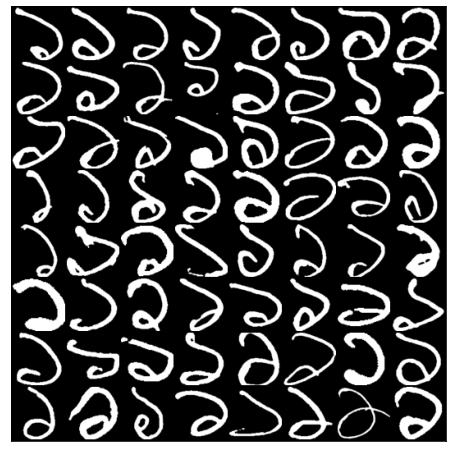}
  \caption{BanglaLekha Isolated Numeric Image of one class.}
\end{subfigure}%
\hfill
\begin{subfigure}{.3\textwidth}
  \centering
  \includegraphics[width=.8\linewidth]{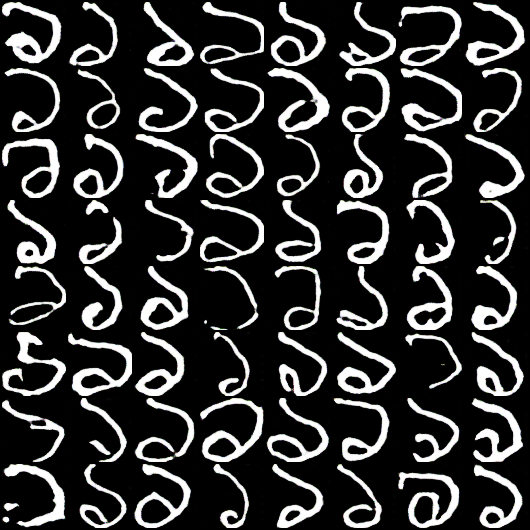}
  \caption{GAN-generated BanglaLekha Isolated Numeric Image of one class after 300 epochs.}
\end{subfigure}

\begin{subfigure}{.3\textwidth}
  \centering
  \includegraphics[width=.8\linewidth]{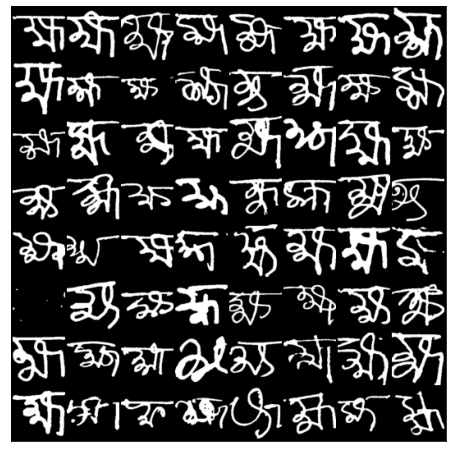}
  \caption{BanglaLekha Isolated Complex Image of one class.}
\end{subfigure}%
\hfill
\begin{subfigure}{.3\textwidth}
  \centering
  \includegraphics[width=.8\linewidth]{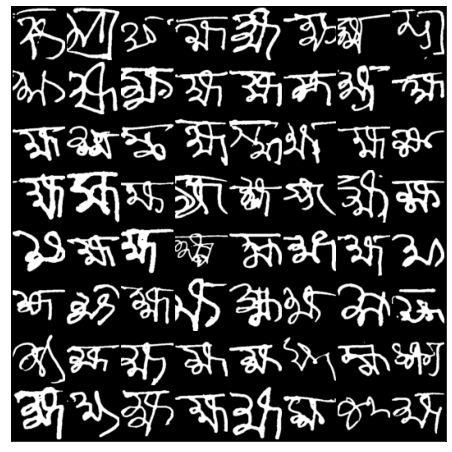}
  \caption{GAN-generated BanglaLekha Isolated Complex Image of one class after 300 epochs.}
\end{subfigure}

\caption{Real and fake sample images of BanglaLekha Isolated Dataset.}
\label{figure4}
\end{figure*}

\begin{table*}[ht]
\centering
\caption{Comparison of accuracy of various models for different epoch sizes on BanglaLekha Isolated Dataset. 
The dataset has been split into 4 groups: V = Vowels (11 classes), C = Consonants (39 classes), 
N = Numerics (10 classes), X = Complex (24 classes), and All = Entire dataset (84 classes).}
\label{tab:commands}

\small
\resizebox{\textwidth}{!}{%
\begin{tabular}{lccccccccccc}
\toprule
\textbf{Model} & \textbf{Epochs} & 
\textbf{Train (V)} & \textbf{Test (V)} &
\textbf{Train (C)} & \textbf{Test (C)} &
\textbf{Train (N)} & \textbf{Test (N)} &
\textbf{Train (X)} & \textbf{Test (X)} &
\textbf{Train (All)} & \textbf{Test (All)} \\
\midrule
3DCNN & 10 & 93.22 & 92.05 & 85.49 & 85.77 & 97.10 & 96.05 & 80.91 & 80.42 & 81.11 & 84.23 \\
3DCNN & 20 & 96.65 & 93.00 & 89.04 & 86.95 & 98.20 & 96.35 & 85.43 & 81.12 & 84.75 & 85.06 \\
3DCNN & 30 & 97.20 & 92.55 & 91.26 & 86.87 & 98.89 & 96.80 & 88.78 & 81.19 & 86.46 & 85.16 \\
3DCNN & 40 & 97.62 & 93.27 & 92.53 & 87.38 & 98.80 & 96.20 & 90.22 & 80.96 & 87.52 & 84.95 \\
3DCNN & 50 & 97.90 & 93.29 & 93.18 & 87.29 & 99.37 & 96.25 & 91.66 & 81.81 & 88.05 & 85.29 \\
\hline
ResNet50 & 10 & 92.34 & 90.12 & 90.26 & 87.56 & 95.78 & 93.17 & 89.12 & 87.57 & 91.16 & 89.81 \\
ResNet50 & 20 & 93.12 & 90.19 & 92.98 & 89.11 & 96.17 & 94.36 & 90.14 & 88.12 & 92.15 & 92.18 \\
ResNet50 & 30 & 95.12 & 92.67 & 93.82 & 90.83 & 96.85 & 94.90 & 90.89 & 88.56 & 93.98 & 92.39 \\
ResNet50 & 40 & 96.56 & 93.23 & 95.38 & 92.91 & 97.12 & 96.12 & 91.15 & 88.72 & 95.47 & 92.67 \\
ResNet50 & 50 & 96.87 & 93.89 & 96.83 & 93.83 & 97.33 & 96.45 & 92.73 & 88.78 & 96.88 & 92.82 \\
\hline
MobileNet & 10 & 93.74 & 90.59 & 89.59 & 90.74 & 98.31 & 98.15 & 87.89 & 86.41 & 90.61 & 89.81 \\
MobileNet & 20 & 94.63 & 91.42 & 91.11 & 92.75 & 98.56 & 98.45 & 88.83 & 89.13 & 91.67 & 91.63 \\
MobileNet & 30 & 96.23 & 95.48 & 91.89 & 92.69 & 98.89 & 98.68 & 89.71 & 89.41 & 92.11 & 91.89 \\
MobileNet & 40 & 96.89 & 96.36 & 92.64 & 92.78 & 99.01 & 98.31 & 89.75 & 87.81 & 92.17 & 92.12 \\
MobileNet & 50 & 98.79 & 97.74 & 93.81 & 93.34 & 99.12 & 98.13 & 91.34 & 90.01 & 93.56 & 93.44 \\
\hline
InceptionV3 & 10 & 98.11 & 96.01 & 94.18 & 94.61 & 98.56 & 98.67 & 91.91 & 92.89 & 91.91 & 91.12 \\
InceptionV3 & 20 & 98.34 & 96.16 & 95.15 & 94.94 & 98.84 & 98.51 & 93.01 & 92.65 & 91.37 & 91.45 \\
InceptionV3 & 30 & 98.45 & 98.14 & 96.19 & 95.36 & 98.96 & 98.61 & 93.43 & 93.41 & 92.81 & 92.98 \\
InceptionV3 & 40 & 99.12 & 97.13 & 96.64 & 95.78 & 99.05 & 98.91 & 94.61 & 93.92 & 93.59 & 93.61 \\
InceptionV3 & 50 & 99.31 & 98.34 & 97.74 & 96.14 & 99.12 & 98.98 & 94.91 & 95.89 & 94.71 & 93.74 \\
\hline
ADA-GAN & 10 & 97.37 & 94.63 & 96.13 & 95.37 & 97.13 & 97.58 & 93.69 & 93.15 & 94.87 & 92.15 \\
ADA-GAN & 20 & 97.78 & 95.62 & 96.39 & 95.48 & 97.31 & 97.89 & 94.38 & 93.75 & 95.02 & 92.18 \\
ADA-GAN & 30 & 98.38 & 97.59 & 96.78 & 96.13 & 98.48 & 98.41 & 95.15 & 94.69 & 95.19 & 93.74 \\
ADA-GAN & 40 & 98.45 & 98.31 & 97.11 & 96.53 & 99.51 & 99.17 & 96.19 & 95.41 & 96.13 & 94.38 \\
ADA-GAN & 50 & 98.71 & 98.45 & 97.56 & 97.67 & 99.73 & 99.24 & 96.65 & 95.91 & 96.54 & 94.94 \\
\bottomrule
\end{tabular}%
}
\end{table*}

\section{Conclusion}
We have proposed ADA-GAN, which consists of three distinct Artificial Neural Networks for classifying highly cursive complex character recognition in a fully supervised environment. We have shown that, after augmenting the training dataset with GAN-generated images with adversarially perturbed noise, our model is getting better results even after the complexity of handwritten characters arises. This work can be extended to generate datasets, especially in the medical, finance, and government domains, with a multi-modal dataset, where the privacy and security of the real data are important and/or costly to generate. Also, this approach can be implemented where the real-world dataset is hard to generate. Moreover, we may extend this work into a conditional GAN where there may be some constrained optimization problem involved that needs to either maximize or minimize any particular parameter.

\end{document}